%% file: arxiv.tex
\documentclass{article}

\PassOptionsToPackage{dvipsnames,table}{xcolor}
\PassOptionsToPackage{numbers, compress}{natbib}

\usepackage[final]{neurips_2025}

\usepackage{amssymb}
\usepackage{pifont}
\usepackage{float}
\usepackage{multirow}
\usepackage{tabularx}
\usepackage{array}
\usepackage{graphicx}
\usepackage{makecell}
\usepackage{booktabs}
\usepackage{subcaption}
\usepackage{hyperref}
\usepackage{url}
\usepackage{tikz}
\usepackage{enumitem}
\usepackage{threeparttable}
\usepackage{wrapfig}
\usepackage[export]{adjustbox}
\input{math_commands.tex}

\newcolumntype{C}[1]{>{\centering\arraybackslash}p{#1}}
\newcolumntype{L}[1]{>{\raggedright\arraybackslash}p{#1}}
\newcolumntype{R}[1]{>{\raggedleft\arraybackslash}p{#1}}

\usepackage[utf8]{inputenc}

\usepackage[most]{tcolorbox}
\tcbuselibrary{listingsutf8}

\newtcblisting{promptbox}[1][]{
  colback=ForestGreen!10,
  colframe=ForestGreen!80,
  listing only,
  listing options={
    basicstyle=\ttfamily\small,
    breaklines=true,
    breakatwhitespace=true
  },
  title={#1},
  boxrule=0.5pt,
  arc=2mm,
  width=\linewidth,     
  left=1mm, right=1mm,  
  top=1mm, bottom=1mm
}
\usepackage[export]{adjustbox}
\newcommand{\bluerectangle}{
    \draw[blue, line width=2pt] (img.south west) rectangle (img.north east);
}
\newcommand{\greenrectangle}{
    \draw[green, line width=2pt] (img.south west) rectangle (img.north east);
}
\newcommand{\yellowrectangle}{
    \draw[yellow, line width=2pt] (img.south west) rectangle (img.north east);
}

\newcommand{\redrectangle}{
    \draw[red, line width=2pt] (img.south west) rectangle (img.north east);
}

\newcommand{\showimagewithoverlay}[3]{
  \begin{tikzpicture}
    \node[inner sep=0pt] (img) {\includegraphics[width=#3]{#1}};
    #2
  \end{tikzpicture}
}

\newcommand{\spotedit}{%
  SpotEdit%
}

\newcommand{\hallucination}{%
  \textit{hallucination}%
}

\newcommand{\Hallucination}{%
  \textit{Hallucination}%
}

\newcommand{\oscore}{%
  \textit{Global Score}%
}

\newcommand{\backscore}{%
  \textit{Background Fidelity}%
}

\newcommand{\objscore}{%
  \textit{Object Fidelity}%
}

\newcommand{\fscore}{%
  \textit{Failure Rate}%
}

\newcommand{\irobscore}{%
  Inp. Robustness%
}

\newcommand{\rrobscore}{%
  Ref. Robustness%
}

\title{\spotedit: Evaluating Visually-Guided Image Editing Methods}       

\author{\textbf{Sara Ghazanfari}$^{1,2*}$,\quad \textbf{Wei-An Lin}$^{2}$,\quad \textbf{Haitong Tian}$^{2}$,\quad \textbf{Ersin Yumer}$^{2}$\vspace{0.5cm}
\\
$^{1}$New York University, US \qquad $^{2}$Adobe Inc.}

\begin{document}

\maketitle

\begin{abstract}
Visually-guided image editing, where edits are conditioned on both visual cues and textual prompts, has emerged as a powerful paradigm for fine-grained, controllable content generation. Although recent generative models have shown remarkable capabilities, existing evaluations remain simple and insufficiently representative of real-world editing challenges. We present \spotedit, a comprehensive benchmark designed to systematically assess visually-guided image editing methods across diverse diffusion, autoregressive, and hybrid generative models, uncovering substantial performance disparities.  To address a critical yet underexplored challenge, our benchmark includes a dedicated component on \hallucination, highlighting how leading models, such as GPT-4o, often hallucinate the existence of a visual cue and erroneously perform the editing task. Our code and benchmark are publicly released at \href{https://github.com/SaraGhazanfari/SpotEdit}{github.com/SaraGhazanfari/SpotEdit}.
\end{abstract}


\section{Introduction}
{%
  \let\thefootnote\relax
  \footnotetext{$^*$Work done during internships at Adobe Inc.}%
}
Visually-guided image editing enables precise, localized manipulation by combining a reference image with textual instructions to guide generative models. Compared to text-only editing, this multimodal approach provides greater control, stronger semantic alignment, and higher spatial precision, making it valuable for applications such as consistent keyframe editing in the area of long-form video editing~\citep{zhang2025adaflow, huang2025generating}.

Despite rapid advances in diffusion-based~\citep{wu2025less, xiao2025omnigen} and autoregressive~\citep{deng2025emerging, sun2024generative} generative methods, rigorous evaluation of visually guided editing remains underexplored. Existing benchmarks~\citep{yang2023paint, li2023dreamedit} focus largely on coarse manipulations, simple object replacements, or single-object scenes (see Fig.~\ref{fig:early-benchmarks}), offering limited insight into the complexities of real-world editing. As a result, current evaluations fail to capture the nuanced challenges of multimodal guidance, hindering fair model comparison and progress.

To close this gap, we introduce \spotedit, a comprehensive benchmark for systematic and fine-grained evaluation of visually guided image editing. \spotedit~is built from diverse real and synthetic video frames, enabling controlled variation in object appearance, position, scale, and context. More specifically, each benchmark instance consists of: a reference image, an input image, and a textual instruction, and a ground-truth target image. Crucially, \spotedit~also includes a dedicated \Hallucination~subset that probes failure cases where objects are missing from either the reference or the input image. This component directly evaluates a model’s robustness to edge cases, testing its ability to avoid spurious insertions while preserving both spatial coherence and semantic fidelity under adverse conditions.

We evaluate leading open- and closed-source models, including OmniGen2~\citep{omniGen2}, BAGEL~\citep{deng2025emerging} and GPT-4o~\footnote{\url{https://openai.com/index/introducing-4o-image-generation/}}
on \spotedit. Results reveal that visually guided editing remains fundamentally challenging: the strongest open-source model achieves only $0.685$ similarity score to ground truth. Moreover, models exhibit complementary strengths and weaknesses, e.g., OmniGen2 adheres closely to visual guidance but disrupts background consistency, while BAGEL preserves context yet struggles with cue interpretation. Strikingly, in \hallucination~cases, even the proprietary GPT-4o performs poorly, hallucinating object presence and executing incorrect edits despite its strong general image editing capabilities.

\section{Related Work}
\label{sec:related-work}
In the following, we present related work on existing visually-guided image editing benchmarks. An extended discussion, including related work on the methods, is provided in App.~\ref{app:extended_related_work}.
\paragraph{Visually-guided image editing benchmarks.}
As a pioneer, Paint by Example\citep{yang2023paint} introduced exemplar-based editing by inpainting image regions from a reference exemplar, focusing on visual similarity and identity preservation. DreamEdit~\citep{li2023dreamedit} extended this to subject-driven editing, manipulating a subject’s appearance or context while preserving identity. As shown in Fig.~\ref{fig:early-benchmarks}, prior benchmarks involve simple scenes, few distractors, and near-identical object poses across reference and edited images. In contrast, \spotedit~targets fine-grained, visually-guided editing in complex scenes with multiple objects, sparse textual descriptions, challenging spatial layouts, and pose variations. Its deterministic design with a ground truth enables objective evaluation and, for the first time, tests \hallucination~to incomplete or missing visual cues.

\section{\spotedit~Benchmark}
\label{sec:spotedit}
In this section, we outline our data generation pipeline, present benchmark statistics, and highlight key characteristics. Additional details are provided in App.~\ref{app:benchmark_details}.

\subsection{Data Generation Pipeline} 
\begin{figure}
    \centering
    \includegraphics[width=\linewidth, trim=0mm 0mm 0mm 0mm]{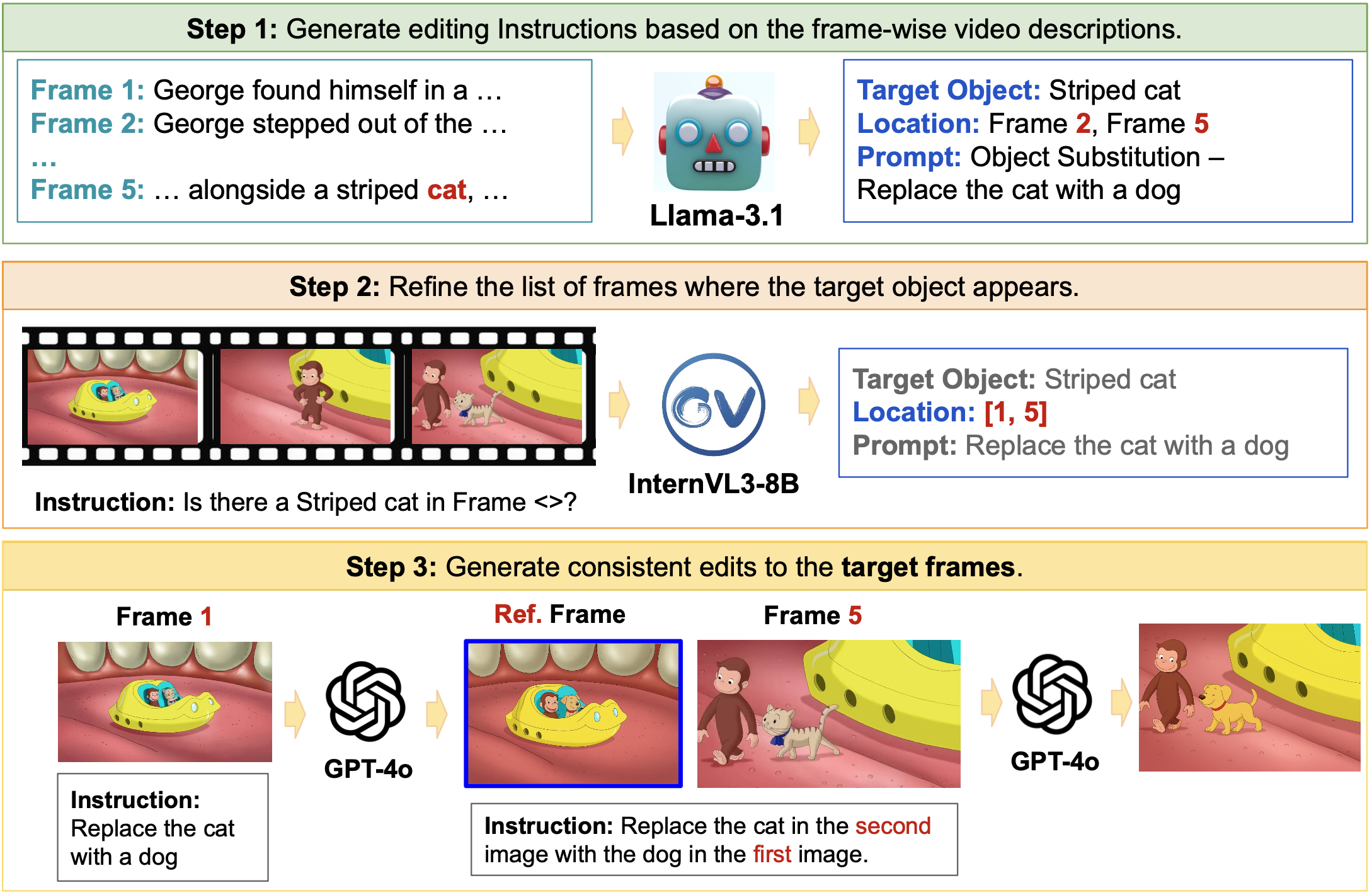}
    \caption{\textbf{SpotEdit data generation pipeline.}
     The pipeline consists of three key stages: 
    (1) generating editing instructions from frame-wise video descriptions, 
    (2) identifying target frames containing the specified object using multimodal queries, and 
    (3) applying consistent edits only to the relevant frames utilizing Visually-guided image editing.
    }
    \label{fig:data-pipe}
    \end{figure}

We construct the SpotEdit benchmark through a structured data generation pipeline that ingests keyframes extracted from video sources. Our data comes from two complementary datasets: (1) the StoryStream dataset~\citep{yang2024seed}, a large-scale, high-resolution synthetic multimodal collection designed for long-form story generation, and (2) real (non-synthetic) videos from NExT-QA ~\citep{xiao2021next}, which add diversity and visual realism. While StoryStream provides keyframes directly, NExT-QA requires preprocessing to extract them, details of which are given in App.~\ref{app:benchmark_details}. Starting with the key-frames, we then utilize our three-stage data generation process, illustrated in Fig.~\ref{fig:data-pipe} and described below, to generate our benchmark:

\textbf{Step 1: Instruction Generation.}  
Both datasets include frame-level captions, which we use to prompt \texttt{Llama-3.1-8B-Instruct}
to produce fine-grained instructions for object-level image editing. The model outputs an editing prompt, the target object, and the corresponding frames. This stage is entirely text-based, relying solely on the narrative captions; no visual inputs are required.

\textbf{Step 2: Frame Localization.}  
To refine the \textit{Location} of the target object, i.e., specific frames where the target object appears, we query \texttt{InternVL3-8B}~\citep{zhu2025internvl3}. 
Each frame is individually evaluated by the model to determine the presence or absence of the target object.

\textbf{Step 3: Consistent Editing.}  
Once the target frames are identified, we perform image editing using the GPT-4o model. More specifically, the first image is edited without any visual guidance. While other frames are edited using the edited version of the first frame as the visual guidance to preserve edit consistency over all keyframes.

We construct benchmark samples by pairing each edit instruction with its corresponding source and edited images. In the standard setting, each sample consists of: a reference frame (providing visual guidance), a source frame (the image to be edited), a textual instruction, and a ground-truth edited frame (the target output). In the \hallucination~setting, we deliberately introduce cases where the target object is absent. Specifically, \rrobscore~samples lack the target object from the reference image, while \irobscore~samples lack it from the source image.
\subsection{Benchmark Statistics \& Key Features} 
Our benchmark contains 500 samples spanning both synthetic and real images. As illustrated in Fig.~\ref{fig:statistics}, approximately 40\% of the samples belong to the \hallucination~setting, while the remaining 60\% correspond to the standard setting.

\spotedit~introduces several distinctive features. First, it explicitly evaluates \hallucination~by including cases where the source object is absent, requiring models to correctly refrain from editing. Second, instead of isolated images, it leverages video keyframes that capture diverse object poses, scales, and lighting conditions, making the task more realistic and challenging. Third, editing instructions are intentionally concise, pushing models to rely on visual grounding for consistency with the reference image. Finally, each task is paired with a nearly unique ground-truth target, allowing precise and unambiguous evaluation, unlike traditional benchmarks. Together, these design choices enable systematic assessment of both standard editing performance and robustness to \hallucination~scenarios.





\section{Experimental Evaluations}
In this section, we evaluate generative models on both \textit{standard} and \hallucination-focused samples from \spotedit, analyzing their strengths and limitations. Further details are provided in App.~\ref{app:evaluation_details}.

\textbf{Evaluation Setup.} 
For our baseline models, we include the most recent models that support visually-guided image editing. Specifically, UNO~\citep{wu2025less} and OmniGen~\citep{xiao2025omnigen} are diffusion-based models; BAGEL~\citep{deng2025emerging} and Emu2~\citep{sun2024generative} are autoregressive generative models; and OmniGen2~\citep{omniGen2} adopts a hybrid architecture that couples an autoregressive text decoder with a diffusion-based image decoder. 
To compute semantic similarity, we use DINOv2~\citep{oquab2023dinov2}
(and later CLIP~\citep{Radford2021LearningTV, cherti2023reproducible} in Tab.~\ref{tab:comparison_scores} of App.~\ref{app:evaluation_details}) to extract image representations, as they have been widely adopted in prior works~\citep{wu2025less, wu2025omnigen2}. Cosine similarity is then computed between the extracted representations, producing scores in the range $0 \leq \text{score} \leq 1$.

\begin{figure}
    \centering
    \begin{subfigure}[b]{1.0\linewidth}
        \includegraphics[width=\linewidth]{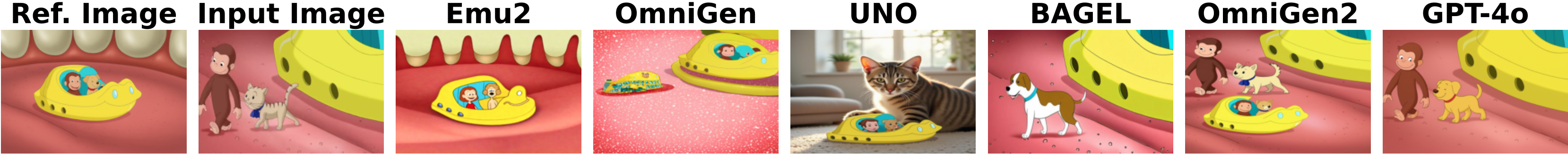}
        \caption{\textit{Instruction}: Replace the striped cat in the second image with the dog in the first image.}
        \label{fig:stan_1}
    \end{subfigure}\\
    \begin{subfigure}[b]{1.0\linewidth}
        \includegraphics[width=\linewidth]{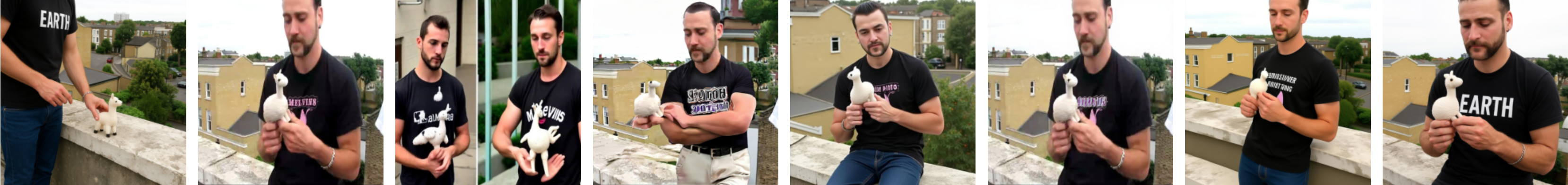}
        \caption{\textit{Instruction}: Match the writing on the man's shirt in the second image with the first image.}
        \label{fig:stan_3}
    \end{subfigure}\\
    \begin{subfigure}[b]{1.0\linewidth}
        \includegraphics[width=\linewidth]{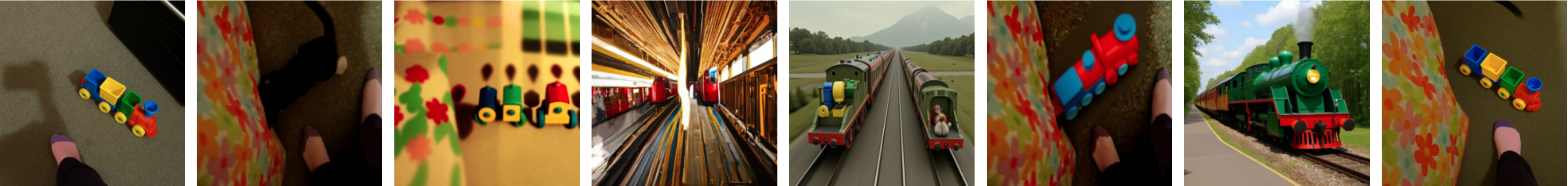}
        \caption{\textit{Instruction}: Replace the cat in the second image with the train in the first image.}
        \label{fig:stan_2}
    \end{subfigure}\\
    \caption{Examples from the \spotedit~\textit{standard} section. Each consists of a reference image, an input image, and an instruction, along with the edited outputs produced by baseline models.}
    \label{fig:standard_category}
    \vspace{-5mm}
\end{figure}

\subsection{Standard Evaluation}

\textbf{Metrics.} 
Qualitative results for both synthetic and real samples in the standard category are presented in Fig.~\ref{fig:standard_category} and Fig.~\ref{fig:more-standard_category}. For quantitative evaluation, we measure performance across three complementary dimensions. The first, \oscore, provides a coarse-grained similarity measure, quantifying the alignment between the edited image and the corresponding ground-truth.
The other two are fine-grained metrics: \objscore~and \backscore. More specifically, \objscore~evaluates whether the model accurately follows the visual guidance from the reference image, preserving the target object’s identity and appearance in the output. \backscore~measures the extent the model maintains the background of the source image while performing the edit.

\textbf{Evaluation Insights.} 
Our results, illustrated in Tab.~\ref{tab:stan_scores}, reveal that visually-guided image editing remains a challenging task even for leading models: the maximum similarity score achieved does not exceed $0.685$.
Moreover, from the \oscore, BAGEL and OmniGen2 emerge as the strongest performers. However, fine-grained analysis uncovers notable differences in their strengths and weaknesses. BAGEL demonstrates strong \backscore~but struggles to follow the visual guidance, leading to lower \objscore. Conversely, OmniGen2 excels at adhering to the reference image’s guidance but exhibits weaknesses in background preservation. These observations are consistent with the qualitative patterns illustrated in Fig.~\ref{fig:standard_category} and Fig.~\ref{fig:more-standard_category}. Finally, real samples appear to pose greater challenges than synthetic ones across almost all models and metrics. Further techincal details on the compuation of these metrics are provided in App.~\ref{app:benchmark_details}. 
\begin{table}[H]
 \vspace{-4mm}
    \centering
    \small
    \tabcolsep=2pt
    \caption{\textbf{Standard evaluation.} BAGEL and OmniGen2 emerge as the strongest performers. Moreover, fine-grained analysis uncovers that BAGEL achieves a strong \backscore~but struggles with visual guidance, while OmniGen2 follows guidance well but preserves backgrounds poorly.}
    \vspace{2mm}
    \label{tab:stan_scores}
    \begin{tabular}{L{20mm}|
    C{15mm}@{\hspace{5pt}}C{15mm}|
    C{15mm}@{\hspace{5pt}}C{15mm}|
    C{15mm}@{\hspace{5pt}}C{15mm}}
        \toprule
        \multirow{2}{*}{Model} & 
        \multicolumn{2}{c}{\oscore} & 
        \multicolumn{2}{c}{\backscore} & 
        \multicolumn{2}{c}{\objscore} \\
        \cmidrule{2-3} \cmidrule{4-5} \cmidrule{6-7}
        & \textbf{Syn} & \textbf{Real} & \textbf{Syn} & \textbf{Real} &  \textbf{Syn} & \textbf{Real} \\
        \midrule
        \textbf{Emu2}     & 0.531           & 0.543             & 0.458             & 0.411          & 0.567          & 0.414 \\
        \textbf{OmniGen}  & 0.380           & 0.252             & 0.340             & 0.283          & 0.391          & 0.223 \\
        \textbf{UNO}      &  0.535          & 0.425             & 0.435             & 0.371          & 0.511           & 0.328 \\
        \textbf{BAGEL}    & \textbf{0.685}  & 0.611             & \textbf{0.797}    & \textbf{0.793} & 0.455          & 0.327 \\
        \textbf{OmniGen2} & 0.670           & \textbf{0.617}    & 0.500             &  0.455         & \textbf{0.719} & \textbf{0.590} \\
        \bottomrule
    \end{tabular}

\end{table}

\subsection{\Hallucination~Evaluation.} 
\label{sec:hall-eval}
\begin{figure}
    \centering
    \begin{subfigure}[b]{1.0\linewidth}
        \includegraphics[width=\linewidth]{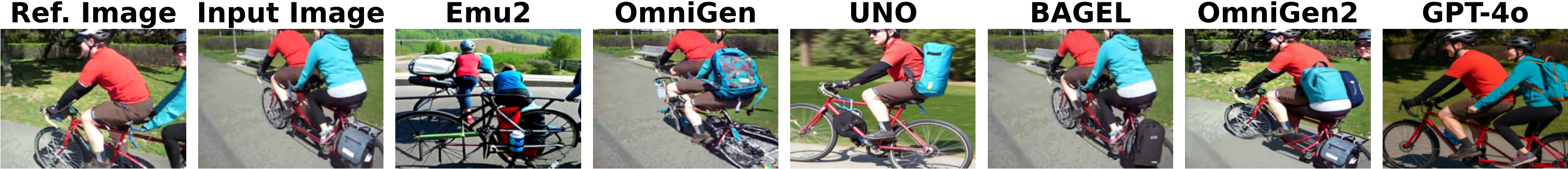}
        \caption{Instruction: Replace the bag in the second image with the backpack shown in the first image.}
        \label{fig:rob_real_1}
    \end{subfigure}\\
    \begin{subfigure}[b]{1.0\linewidth}
        \includegraphics[width=\linewidth]{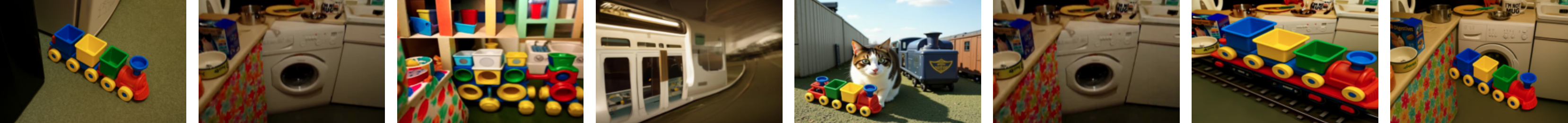}
        \caption{Instruction: Replace the cat in the second image with the train in the first image.}
        \label{fig:rob_real_2}
    \end{subfigure}\\
    \begin{subfigure}[b]{1.0\linewidth}
        \includegraphics[width=\linewidth]{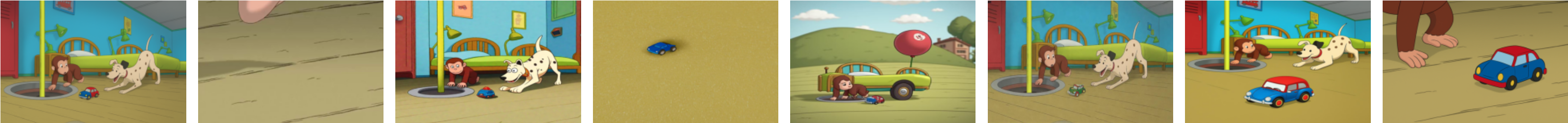}
        \caption{Instruction: Replace the red ball in the second image with the toy car in the first image.}
        \label{fig:rob_syn_3}
    \end{subfigure}\\

    \caption{Examples from the \spotedit~\hallucination~section.}
    \label{fig:rob_category}
    \vspace{-5mm}
\end{figure}

In our benchmark, we include a dedicated section for \hallucination~evaluation, which tests scenarios where either the reference image or the input image does not contain the object specified in the instruction. Representative examples from this section are shown in Fig.~\ref{fig:rob_category} and Fig.~\ref{fig:more-rob_category}. Specifically, Fig.~\ref{fig:rob_real_1} shows a \rrobscore, while Fig.~\ref{fig:rob_real_2} illustrates a \irobscore. In such situations, we expect the model to recognize the absence of the required object and output the unmodified input image.

\textbf{Evaluation Insights.} The \hallucination~results reported in Tab.~\ref{tab:robustness_scores} indicate that this setting is substantially more challenging: all open-source models, except BAGEL, exhibit a marked performance drop relative to their \oscore~in the standard evaluation (Table\ref{tab:stan_scores}). We also evaluate GPT-4o in this setting and observe that, despite its strong general capabilities in image editing, it introduces unintended modifications to the output image, which leads to a low similarity score.

A closer examination of the qualitative examples in Fig.~\ref{fig:standard_category} and Fig.~\ref{fig:more-standard_category} confirms that the models indeed produce hallucinations during the editing process. 
To better quantify hallucination-induced failures, we employ InternVL3-8B\citep{zhu2025internvl3} as a binary classifier to assess the presence of the target object in the generated image. As shown in Tab.~\ref{tab:robustness_scores}, the failure rates are alarmingly high, with GPT-4o emerging as the most vulnerable model in 2 out of the 4 \fscore~evaluations.

Despite the overall poor performance across models, BAGEL outperforms all others on 6 out of 8 evaluation metrics. First, from the standard evaluation, we observe that BAGEL demonstrates strong background-preservation capabilities, which helps maintain a high \oscore. Moreover, BAGEL’s unified design for both generation and understanding tasks equips it with stronger visual understanding capabilities for handling such challenging scenarios.

\begin{table}[htbp]
\vspace{-2mm}
\caption{\textbf{\Hallucination~evaluation.} While GPT-4o performs poorly, emerging as the most vulnerable model in 2 out of the 4 \fscore~evaluations, BAGEL shows strong robustness.}
\label{tab:robustness_scores}
\centering
\resizebox{\textwidth}{!}{%
\begin{tabular}{L{20mm}|*{1}{C{10mm}C{10mm}}|*{1}{C{10mm}C{10mm}}|*{1}{C{10mm}C{10mm}}|*{1}{C{10mm}C{10mm}}}
\toprule
 \multirow{3}{*}{Model}  & \multicolumn{4}{c}{\cellcolor[HTML]{FAF5A5}\textbf{\irobscore}} & \multicolumn{4}{c}{\cellcolor[HTML]{FCDCB6}\textbf{\rrobscore}} \\
 \cmidrule{2-9}
 & \multicolumn{2}{c}{\oscore~$\uparrow$} & \multicolumn{2}{c}{\fscore~(\%)$\downarrow$} & \multicolumn{2}{c}{\oscore~$\uparrow$} & \multicolumn{2}{c}{\fscore~(\%)$\downarrow$} \\ 
 & \textbf{Syn} & \textbf{Real} & \textbf{Syn} & \textbf{Real} & \textbf{Syn} & \textbf{Real} & \textbf{Syn} & \textbf{Real}  \\
\midrule
\textbf{GPT-4o}   & 0.710  & 0.550  & 81.2 & 91.7 & 0.745 & 0.599 & 75.5 & 72.0  \\ 
\midrule
\textbf{Emu2} & 0.440  & 0.338 & 82.7 & 84.0 & 0.544 & 0.402 & \textbf{54.0} & 72.5 \\ 
\textbf{OmniGen}  & 0.285 & 0.260  & 67.3 & 62.0 & 0.371 & 0.372 &  70.0 & 68.6  \\ 
\textbf{UNO}      & 0.383 & 0.350  & 82.7 & 76.0 & 0.419 & 0.433 &  60.0 & 62.7 \\
\textbf{BAGEL}    & \textbf{0.867} & \textbf{0.845} & \textbf{61.5} & \textbf{56.0} & \textbf{0.880} & \textbf{0.735} & 70.0 & 74.5    \\ 
\textbf{OmniGen2} & 0.466 & 0.331 & 84.6 & 88.0 & 0.577 & 0.513 & 64.0 & \textbf{56.9}\\ 
\bottomrule
\end{tabular}}%
\vspace{-2mm}
\end{table}

\section{Conclusion}
We presented \spotedit, a benchmark that brings realistic, fine-grained, and \hallucination-focused evaluation to visually-guided image editing. Testing leading generative models reveals that while some excel in object fidelity or background preservation, none, even the leading closed-source model GPT-4o, achieve consistent performance across all scenarios, especially when visual cues are incomplete. By exposing these gaps, \spotedit~provides a clear path for advancing models that remain accurate and reliable under real-world editing challenges.

\bibliographystyle{ieeenat_fullname}
\bibliography{neurips_2025}
\appendix

\clearpage
\newpage
\section{Extended Related Work}
\label{app:extended_related_work}
\paragraph{Image editing.} Recent advances in image editing have introduced models that take an input image along with a textual edit instruction and produce an edited version of the image~\citep{brooks2023instructpix2pix, yu2025anyedit, liu2025step1x, zhang2025context, lin2025uniworld, wei2024omniedit}. However, to provide more concrete guides for image editing one can provide visual guides as wellas textual guide which lead to a more nuanced task called visually-guided  image editing. These visual cues help convey richer, more precise editing intent, especially when consistency across multiple frames is required. For example, when editing keyframes in a video, maintaining temporal and stylistic coherence across frames is crucial—something that textual instructions alone cannot achieve. This motivates the use of visually-guided  methods, which incorporate reference visuals to better preserve detail and ensure consistency. As expected the methods mentioned before are not capable of performing these tasks. A few recent models have been proposed to address this more complex task.

\paragraph{Visually-guided image editing methods.} (visually-guided) Image editing models typically fall into three categories: diffusion models~\citep{xiao2025omnigen,wu2025less}, autoregressive models~\citep{sun2024generative,sun2024x,deng2025emerging}, and hybrid models~\citep{omniGen2, pan2025transfer} that combine both approaches. Diffusion models including UNO~\citep{wu2025less} and OmniGen~\citep{xiao2025omnigen} use iterative denoising to produce high-quality images, with UNO focusing on visually-guided controllability and OmniGen unifying tasks like text-to-image generation and editing. Autoregressive models including BAGEL~\citep{deng2025emerging}, X‑Prompt~\citep{sun2024x} and Emu2~\citep{sun2024generative} generate content sequentially via next-token prediction.
Bridging the two paradigms, OmniGen2~\citep{omniGen2} adopts a hybrid architecture that couples an autoregressive text decoder with a diffusion-based image decoder, allowing for specialized yet coordinated multimodal generation.  
More recently, a new paradigm has emerged that unifies visual understanding~\citep{ghazanfari2024emma,ghazanfari2025chain} with generation, with BAGEL~\citep{deng2025emerging} being a representative example of this category.
\begin{figure}[H]
    \begin{minipage}{0.45\linewidth}
        \centering
        Paint by Example~\citep{yang2023paint}
    \end{minipage}\hfill
    \begin{minipage}{0.45\linewidth}
        \centering
        DreamEdit~\citep{li2023dreamedit}
    \end{minipage}
    \vspace{2mm}
    \begin{minipage}{\linewidth}
        
        \showimagewithoverlay{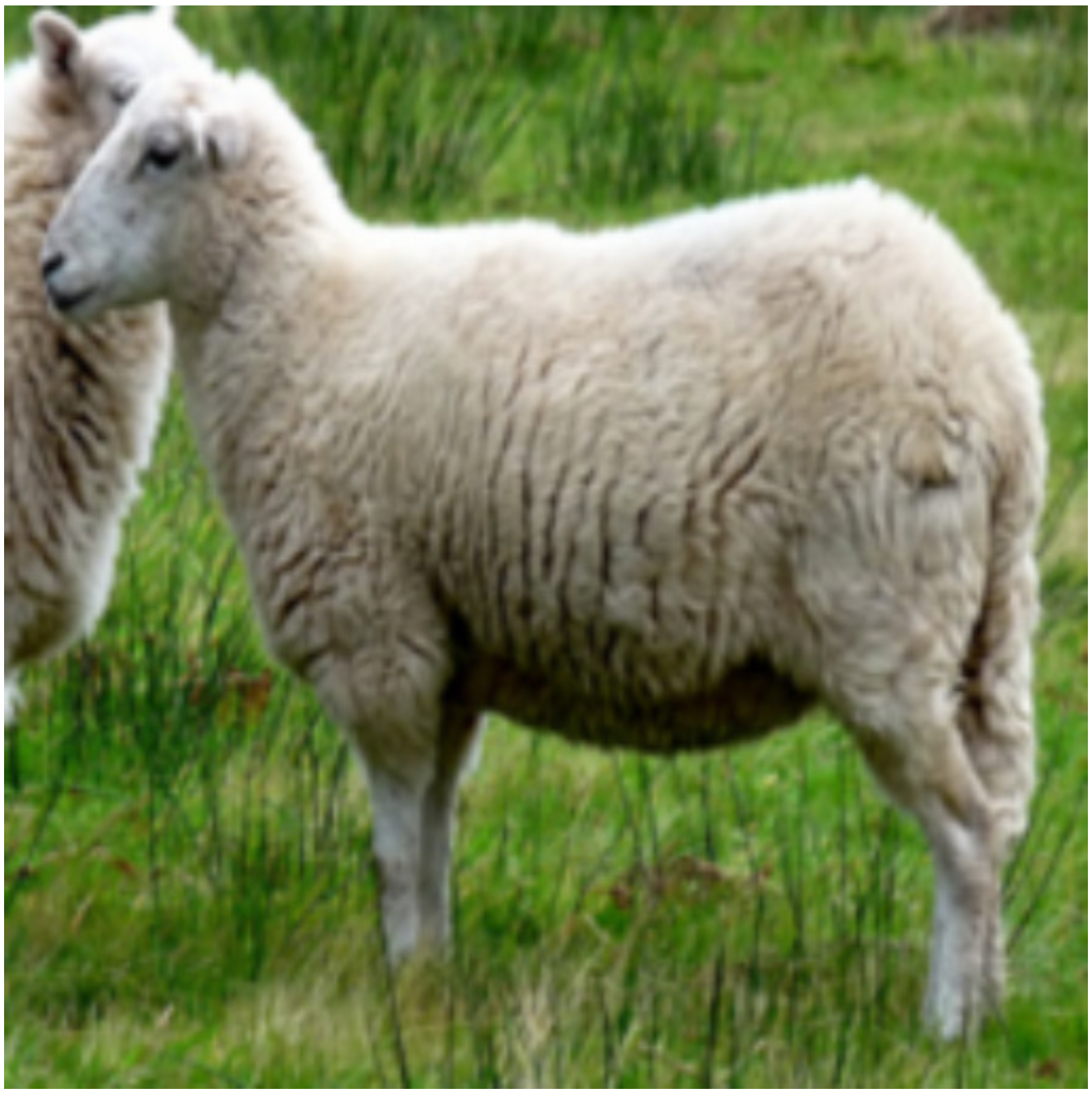}{\bluerectangle}{0.13\linewidth}
        \showimagewithoverlay{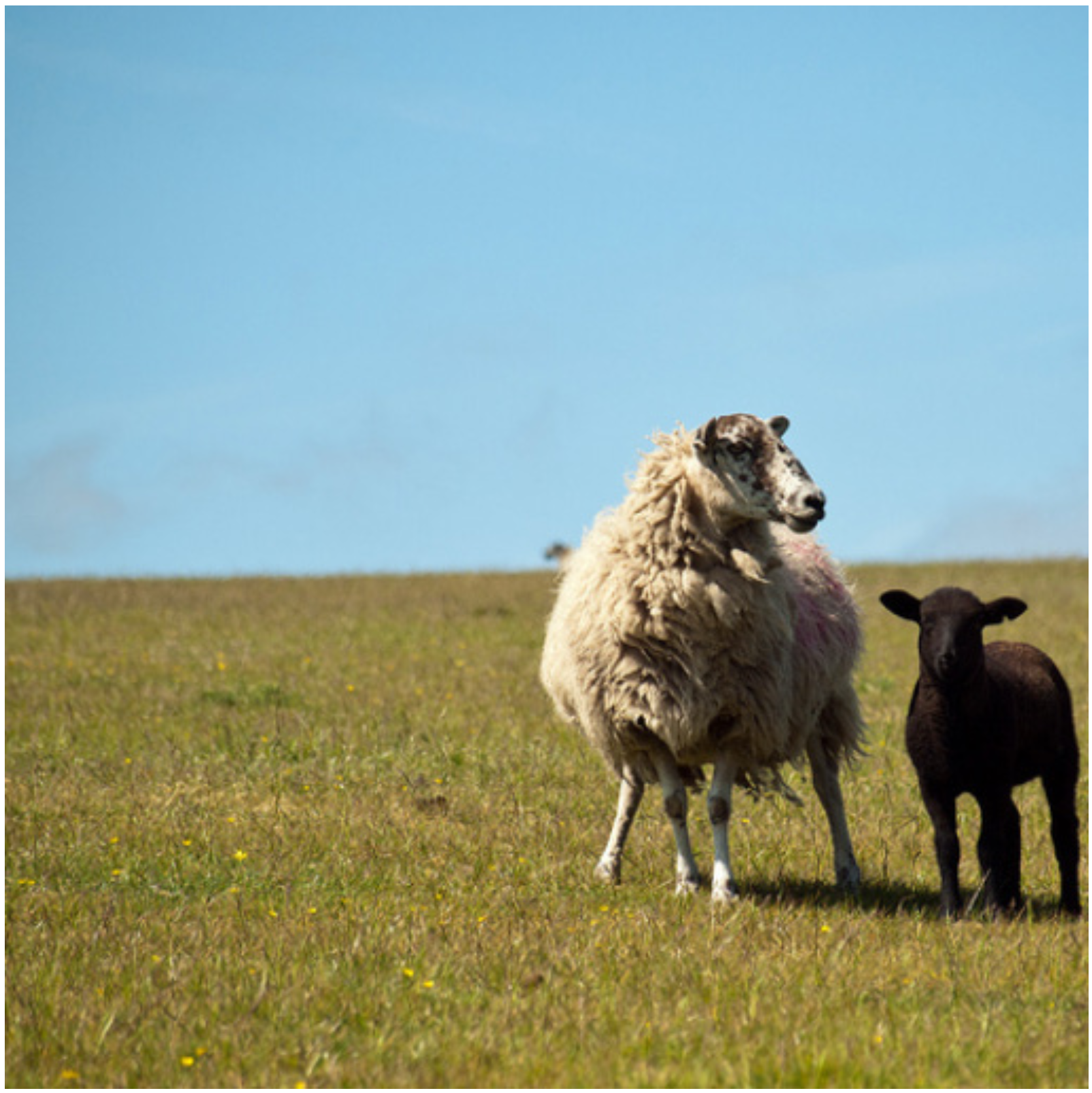}{\yellowrectangle}{0.13\linewidth}
        \showimagewithoverlay{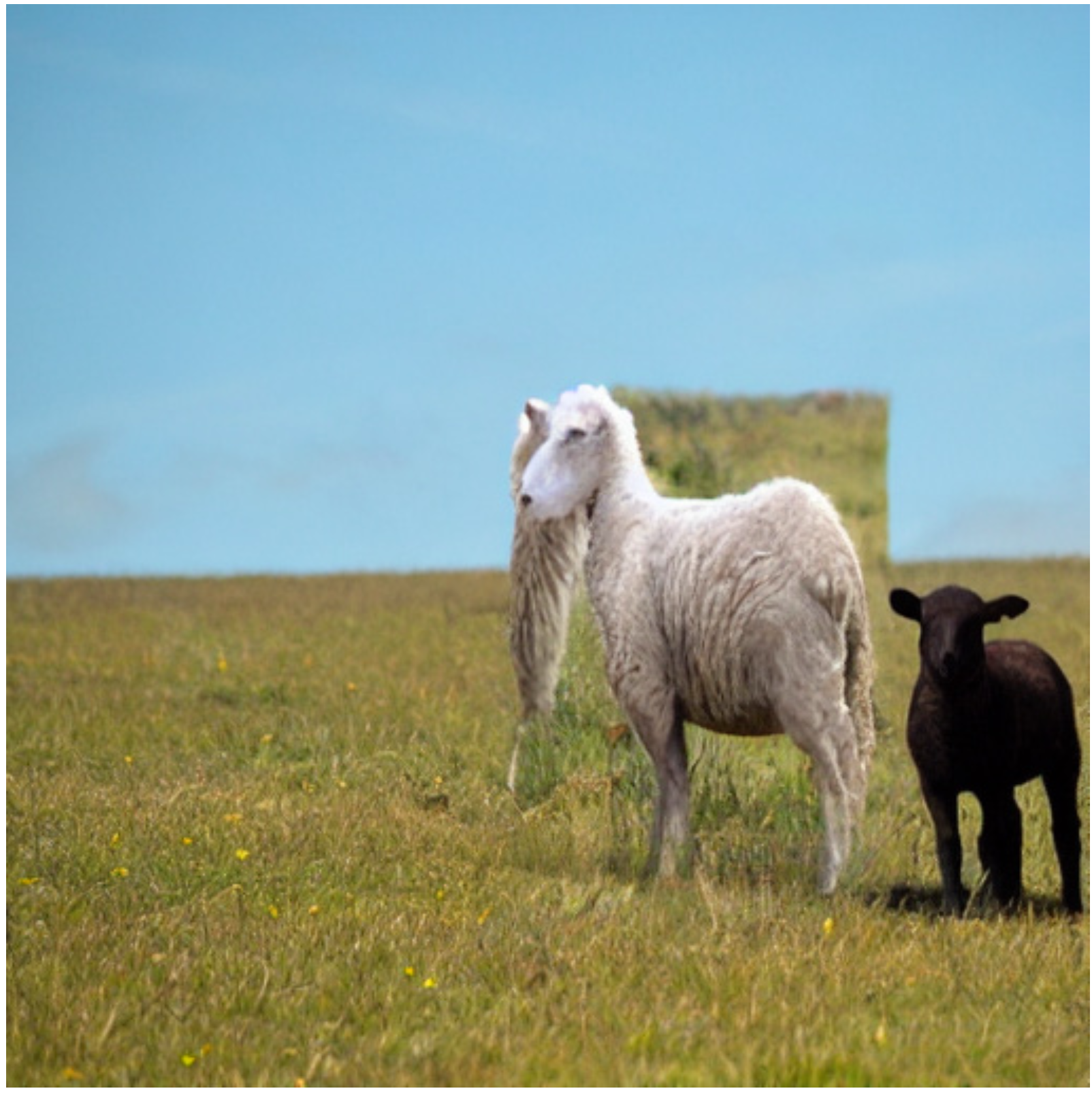}{\greenrectangle}{0.13\linewidth}\hfill
        \showimagewithoverlay{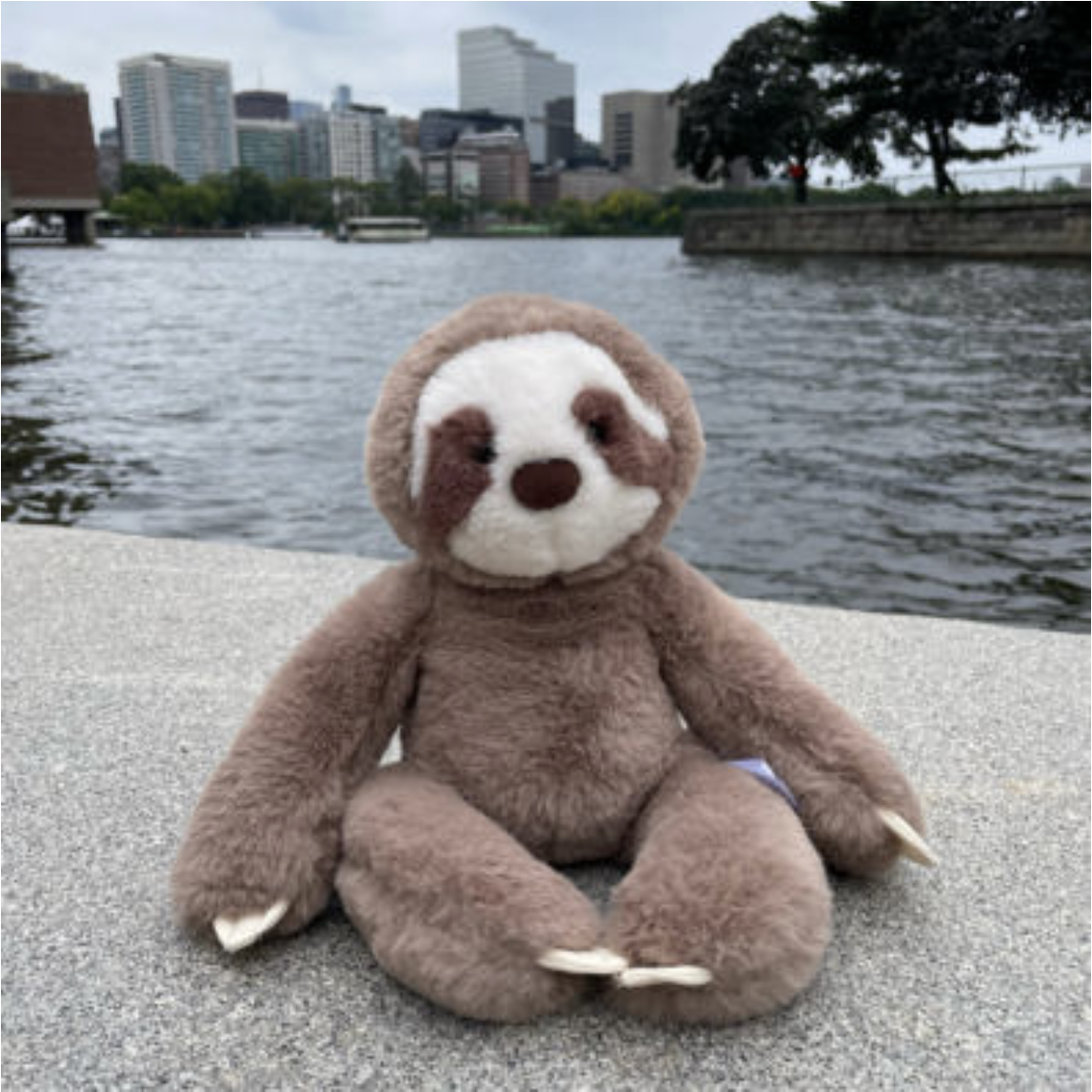}{\bluerectangle}{0.13\linewidth}
        \showimagewithoverlay{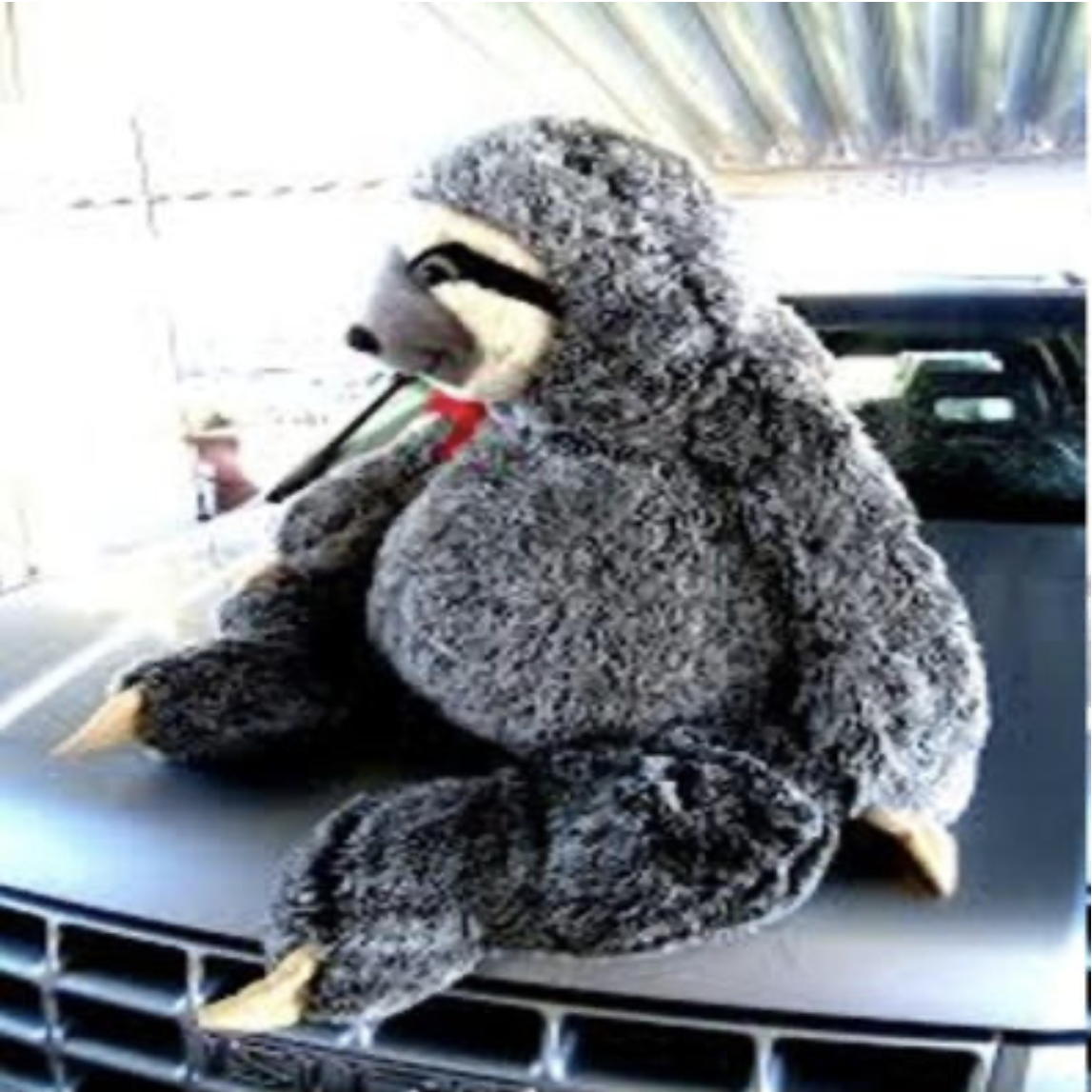}{\yellowrectangle}{0.13\linewidth}
        \showimagewithoverlay{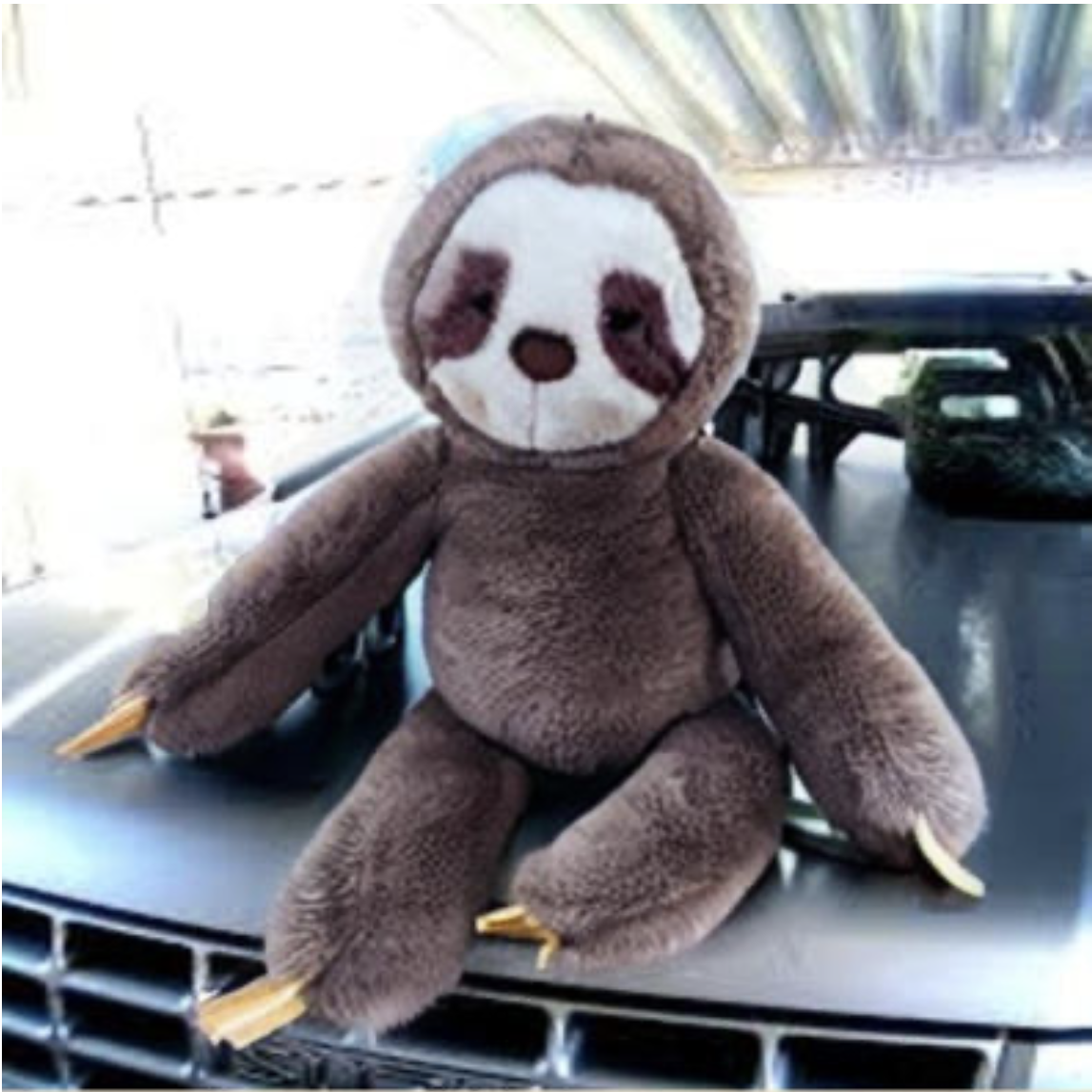}{\greenrectangle}{0.13\linewidth}
    \end{minipage} 
    \caption{Early Visually-guided image editing benchmarks.}
    \label{fig:early-benchmarks}
\end{figure}
\paragraph{Visually-guided image editing benchmarks.}
Paint by Example\citep{yang2023paint} introduced exemplar-based editing by inpainting image regions from a reference exemplar, focusing on visual similarity and identity preservation. DreamEdit~\citep{li2023dreamedit} extended this to subject-driven editing, manipulating a subject’s appearance or context while preserving identity. As shown in Fig.~\ref{fig:early-benchmarks}, prior benchmarks involve simple scenes, few distractors, and near-identical object poses across reference and edited images. 

Unlike previous benchmarks, our work focuses on fine-grained visually-guided image editing, where both the reference and input images contain multiple, distinct objects and exhibit complex spatial structures. Editing models must first detect and identify the target object in the reference image, then accurately locate and modify the corresponding object in the input image; all while preserving the object's identity and maintaining the overall visual coherence of the scene. Another key distinction of our benchmark is its use of minimal prompts: unlike prior datasets that describe the expected final image in detail, our approach relies heavily on visual cues from the reference image, requiring models to reason more deeply about visual context. This makes our benchmark particularly challenging, as it demands both compositional consistency and semantic reasoning across multiple images. Furthermore, in contrast to other editing tasks that allow for open-ended outputs, our benchmark is deterministic—each editing task has a unique, ground-truth result. This enables precise and objective evaluation of model performance.

A parallel line of research~\citep{wu2025omnigen2, liang2025vodiff, kim2025orida} has focused on image composition tasks, where a new image is synthesized by combining elements from multiple context images. Unlike visually-guided editing, these works do not involve editing a specific input image while preserving its structure and style; instead, they aim to generate a novel composition based on the contextual images alone.

\section{\spotedit~Details}
\label{app:benchmark_details}



In this section, we first explain the preprocessing step i.e., extracting the keyframes from the real videos video. Moreover, we provide the prompts used in each step of the pipeline to query the generative model.

\paragraph{Keyframe Extraction} We construct the SpotEdit benchmark using a structured data generation pipeline, as illustrated in Fig.~\ref{fig:data-pipe}. The pipeline takes as input keyframes extracted from various video sources. For synthetic datasets such as StoryStream\citep{yang2024seed}, keyframes are provided directly. For the ShareGPT-4\citep{xiao2021next} dataset, we employ \textsc{CLIP-ViT-H-14}~\citep{Radford2021LearningTV, cherti2023reproducible} combined with cosine similarity to measure the similarity between consecutive frames. Specifically, the selection process begins with the first frame and progressively includes subsequent frames that have a similarity score less than 0.8 compared to the most recently selected keyframe. 

\paragraph{Data generation pipeline.} 
In the first step of the pipeline, the frame captions are provided to the \texttt{Llama-3.1-8B-Instruct}\footnote{\url{https://huggingface.co/meta-llama/Llama-3.1-8B-Instruct}}, which generates the corresponding edit instructions. The prompt used in this step is shown in Fig.~\ref{fig:step-1-prompt}. For Steps 2 and 3, the instructions are shown in Fig.~\ref{fig:data-pipe}.

\begin{figure}
    \centering
    \begin{tabular}{m{2.5cm}
                >{\centering\arraybackslash}m{0.2\linewidth}
                >{\centering\arraybackslash}m{0.2\linewidth}
                >{\centering\arraybackslash}m{0.02\linewidth}
                >{\centering\arraybackslash}m{0.2\linewidth}}
        &  Frame 1 & Frame 2 &  & Frame 5 \\
       \centering\textbf{Source Frames} &
        \showimagewithoverlay{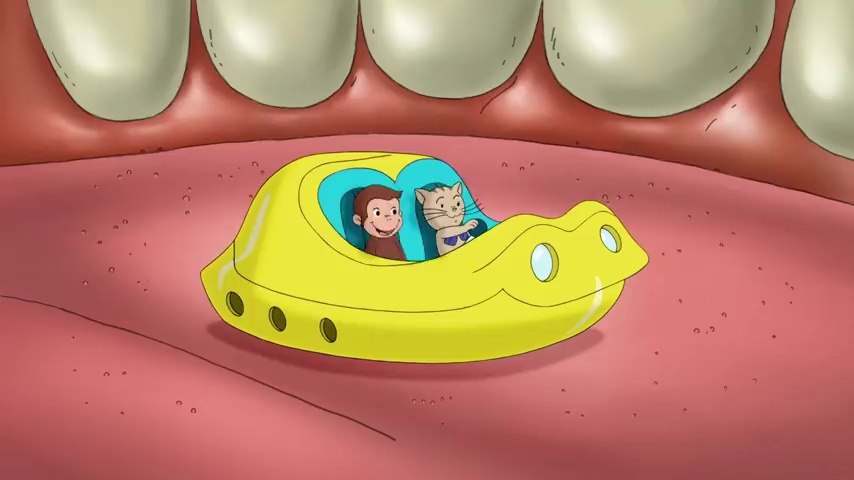}{\bluerectangle}{1.0\linewidth} &
        \showimagewithoverlay{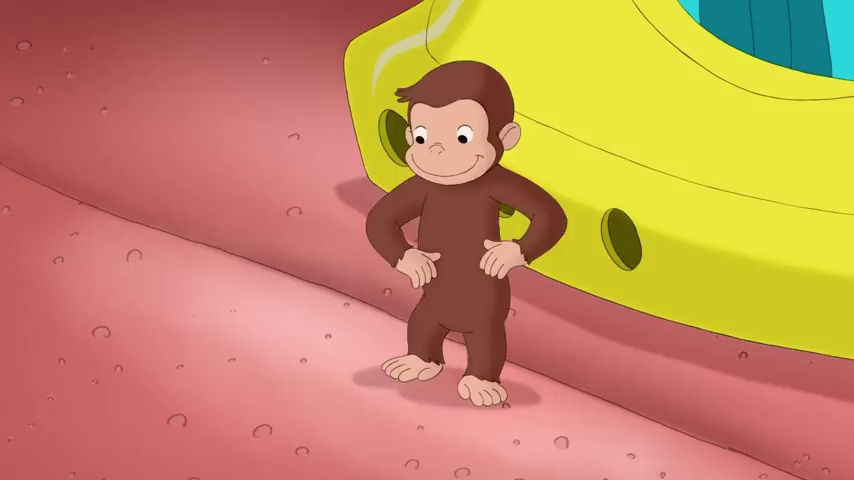}{\redrectangle}{1.0\linewidth}  & ... &
        \showimagewithoverlay{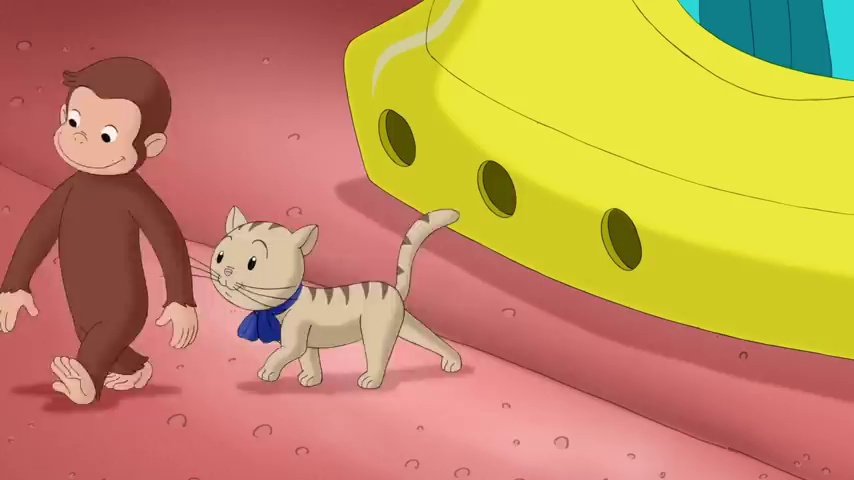}{\bluerectangle}{1.0\linewidth}  \\[4pt]
        \centering\textbf{Edited Frames} &
        \includegraphics[width=1.0\linewidth, height=1.5cm]{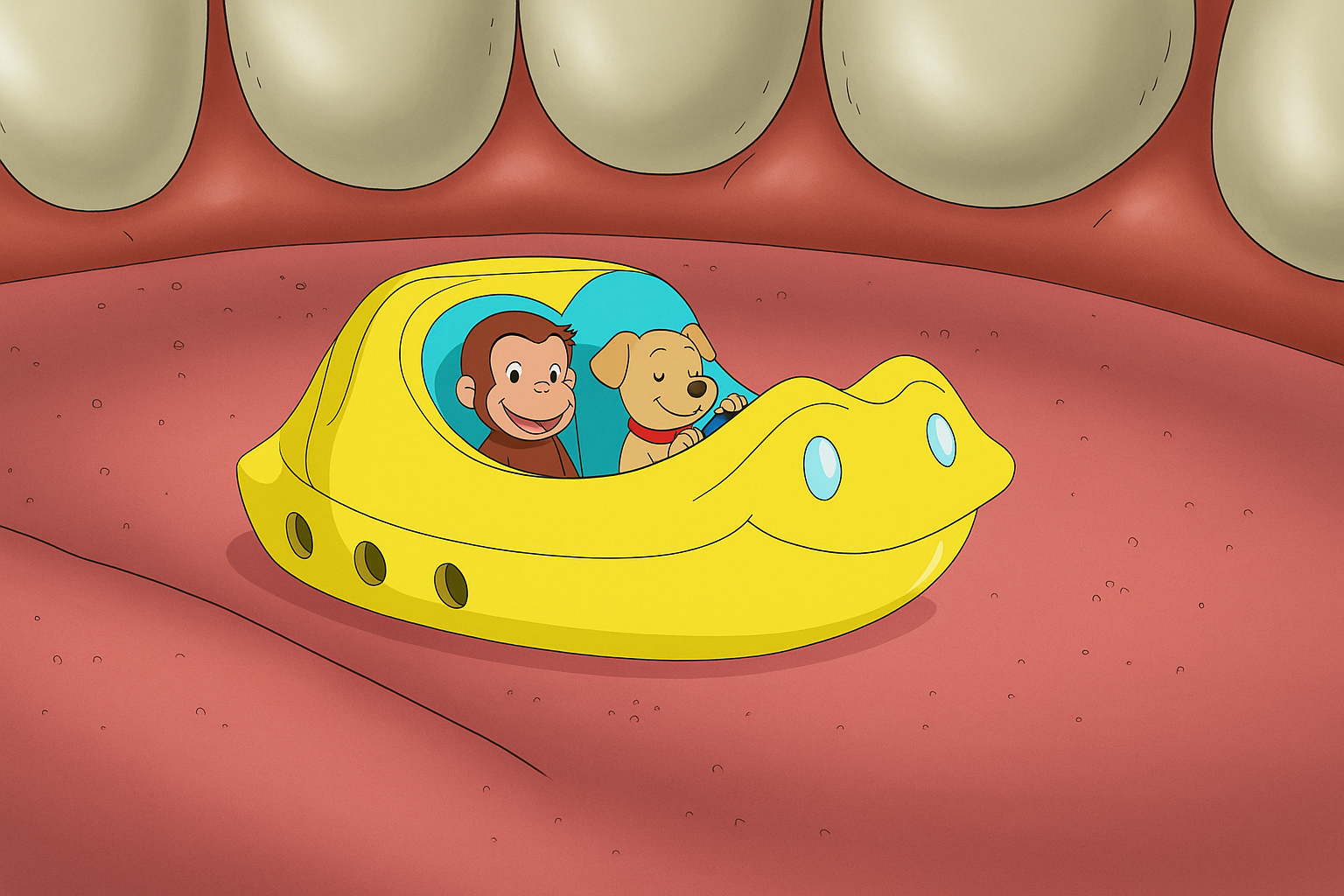} &
        \includegraphics[width=1.0\linewidth]{figs/spotedit/2_org.jpg} & ... &
        \includegraphics[width=1.0\linewidth, height=1.5cm]{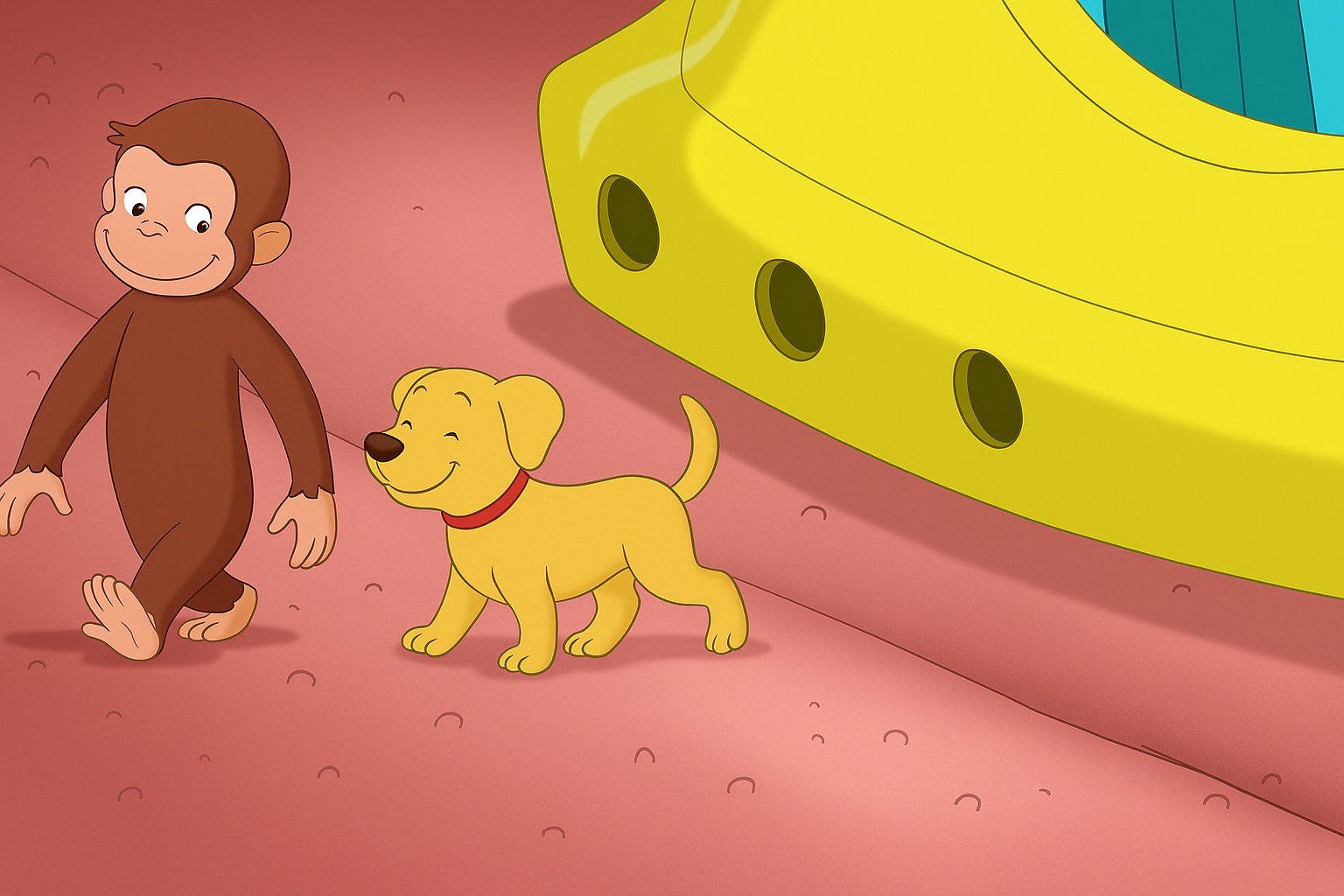} \\[10pt]
        &\multicolumn{4}{l}{\textbf{(a)} Instruction: Replace the cat with a dog.} \\
        \cmidrule{2-5}
       \centering\textbf{Source Frames} &
        \showimagewithoverlay{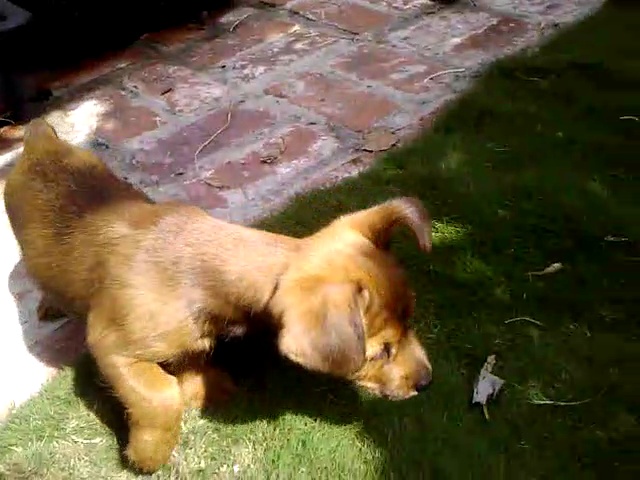}{\redrectangle}{1.0\linewidth} &
        \showimagewithoverlay{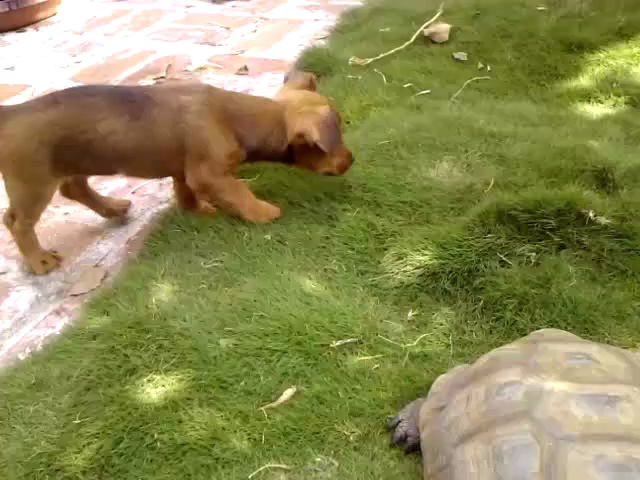}{\bluerectangle}{1.0\linewidth}  & ... &
        \showimagewithoverlay{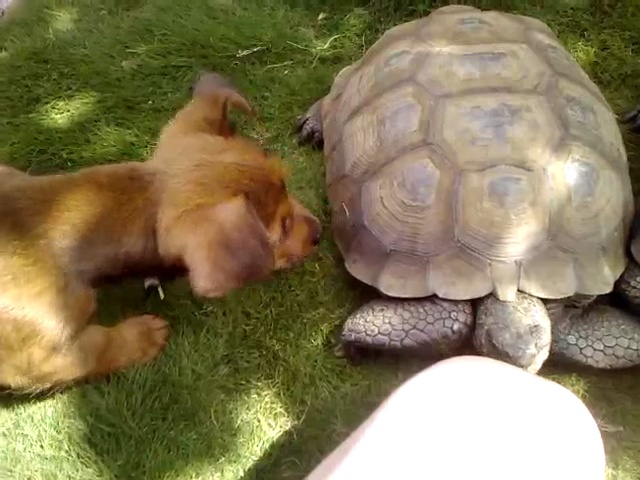}{\bluerectangle}{1.0\linewidth}  \\[4pt]
        \centering\textbf{Edited Frames} &
        \includegraphics[width=1.0\linewidth]{figs/spotedit/real/1.jpg} &
        \includegraphics[width=1.0\linewidth, height=2cm]{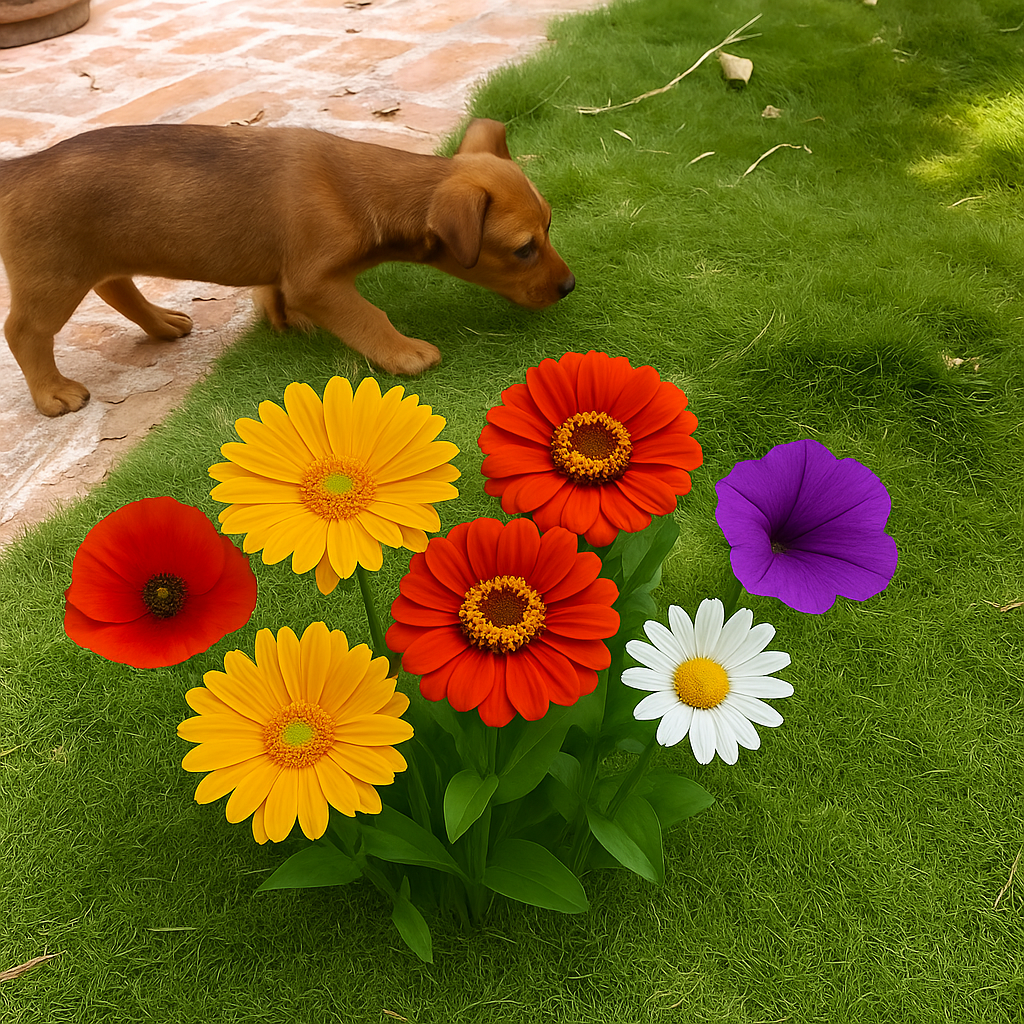} & ... &
        \includegraphics[width=1.0\linewidth, height=2cm]{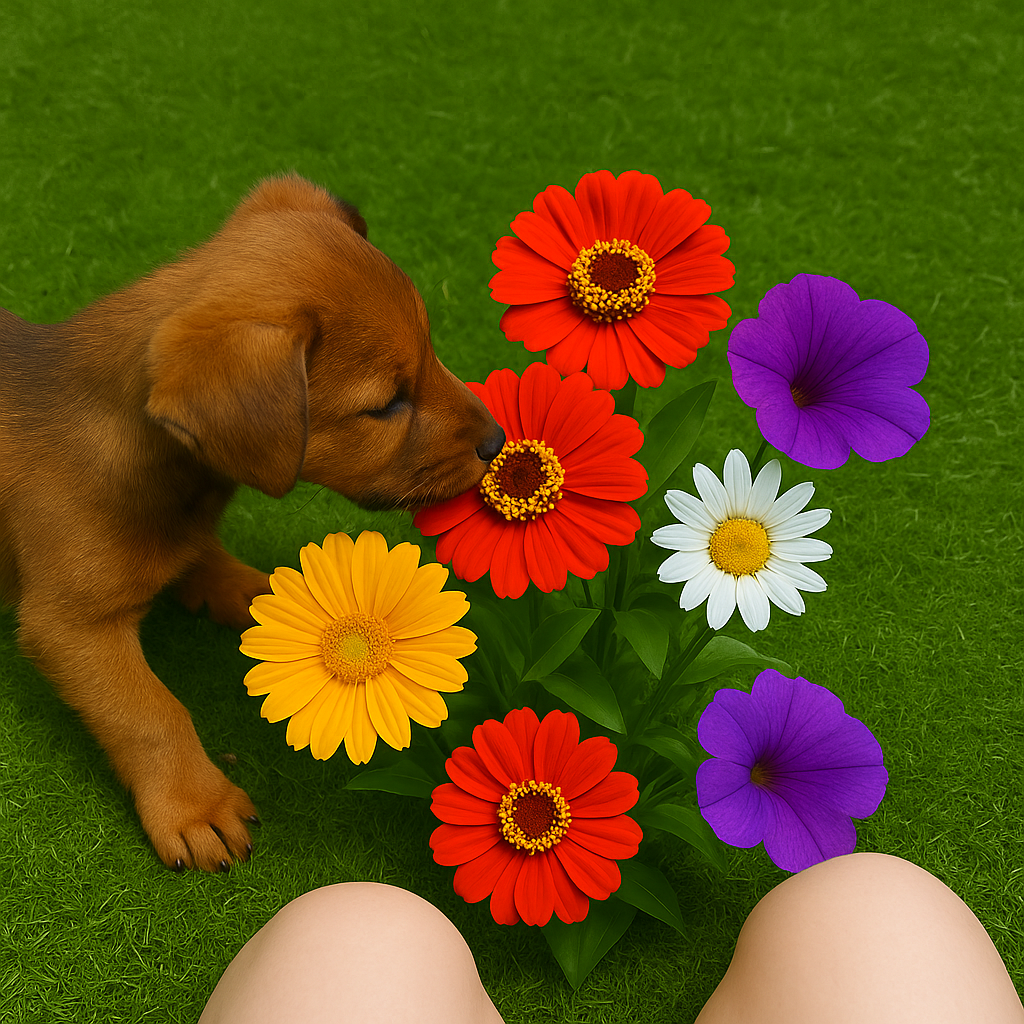} \\
        &\multicolumn{4}{l}{\textbf{(b)} Instruction: Replace the tortoise with a bunch of flowers.} \\
    \end{tabular}
        \caption{Examples generated by our data pipeline. Target frames (blue) are used to construct the \textit{standard} section of the benchmark, while untargeted frames (red) are used to construct the \hallucination~section of our benchmark.}
    \label{fig:bench-detailed-example}
\end{figure}

After running the data generation pipeline, we obtain consistent edits applied to the target frames, as illustrated in Fig.~\ref{fig:bench-detailed-example}. The frames containing the specified object (e.g., Frame 1 and Frame 5) form the standard samples. The remaining frames (e.g., Frame 2), where the specified object is absent, form the \hallucination~samples. We further divide the \hallucination~samples into two subcategories:
\begin{itemize}[itemsep=0mm, leftmargin=*]
\item \irobscore: The reference image contains the target object (e.g., Edited Frame 1 or 5), but the input image does not contain the source object (e.g., Frame 2).
\item \rrobscore: The reference image does not contain the target object (e.g., Frame 2), but the input image contains the source object (e.g., Frame 1 or Frame 5).
\end{itemize}

\begin{figure}[H]
    \centering
\begin{promptbox}[Step 1 Prompt]
You will be provided with a detailed description of a video, including frame-by-frame events.

Your task is to generate a clear, concise, and specific editing instructions based on this description. 
Focus exclusively on the following types of edits:

- Object Modification (e.g., adjusting color, shape, size, or type)
- Object Removal
- Object Substitution 

It is essential that you clearly identify the target object, including any distinguishing features, location, or context, so that it can be reliably recognized and edited.
Response should contain:
* Target Object: 
* Location: 
* Instruction:
\end{promptbox}
\vspace{-2mm}
\caption{\textbf{Step 1 prompt.}.}
    \label{fig:step-1-prompt}
\end{figure}

\begin{figure}[H]
    \centering
    \includegraphics[width=0.5\linewidth, trim=0mm 0mm 0mm 15mm]{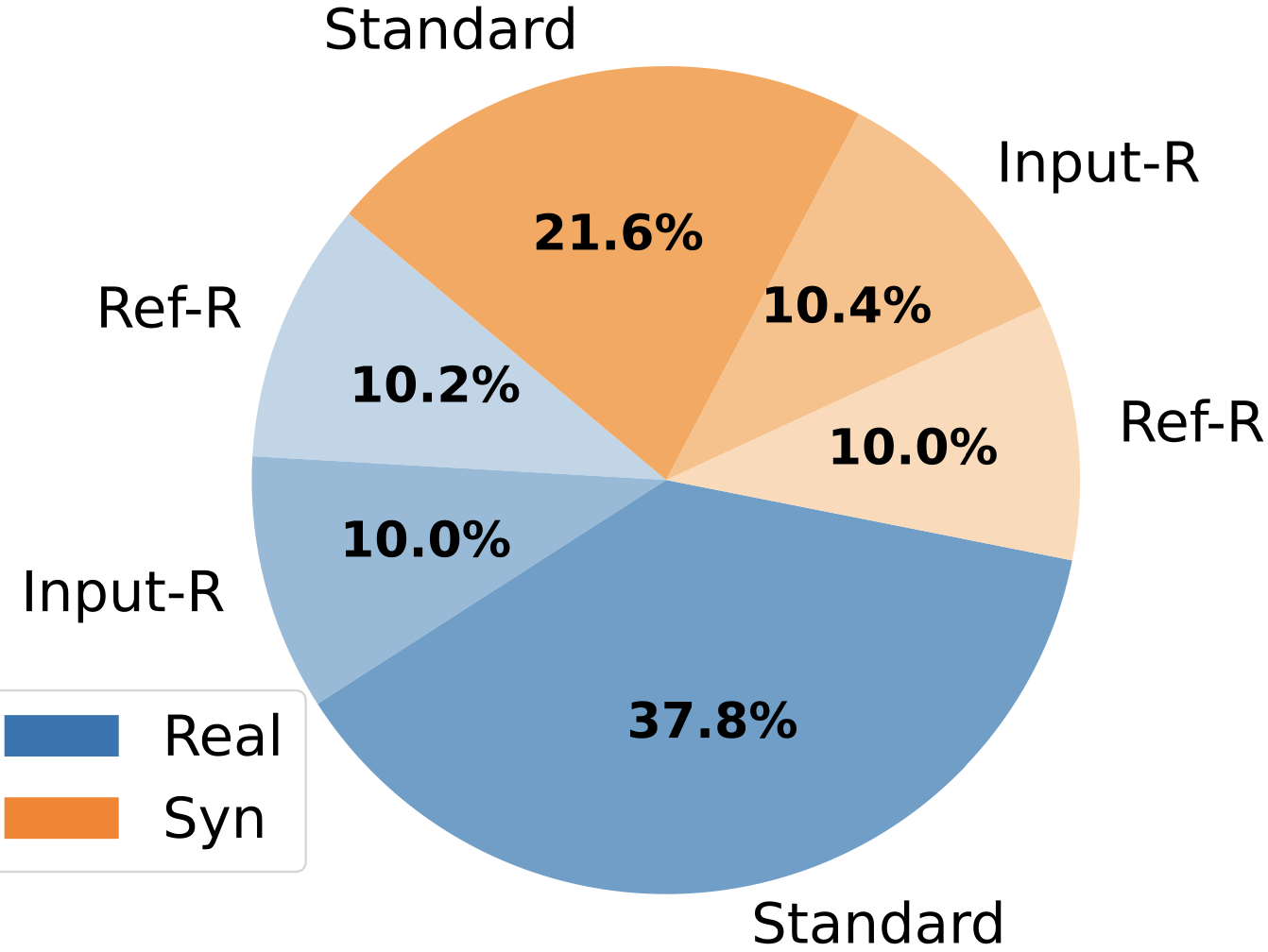}
    \caption{Statistics of the \spotedit~benchmark across 500 samples.}
    \label{fig:statistics}
\end{figure}

\begin{figure}[H]
    \centering
    \begin{subfigure}[b]{1.0\linewidth}
        \includegraphics[width=\linewidth]{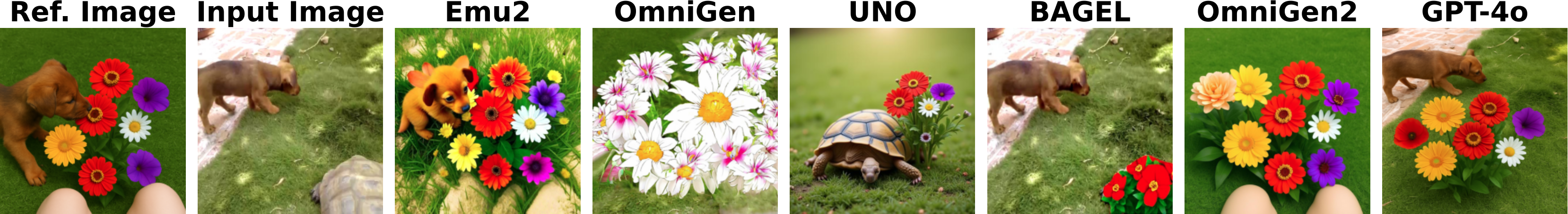}
        \caption{Instruction: Replace the tortoise in the second image with the flowers shown in the first image.}
        \label{fig:app_real_1}
    \end{subfigure}
    \begin{subfigure}[b]{1.0\linewidth}
        \includegraphics[width=\linewidth]{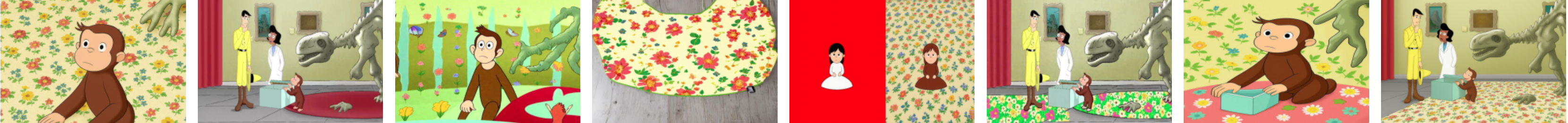}
        \caption{Instruction: Replace the red surface in the second image with a flowery surface in the first image.}
        \label{fig:app_syn_2}
    \end{subfigure}\\
    \begin{subfigure}[b]{1.0\linewidth}
        \includegraphics[width=\linewidth]{figs/benchmarks/spotedit/real_2.png}
        \caption{Instruction: Match the writing on the man's shirt in the second image with the first image.}
        \label{fig:app_real_2}
    \end{subfigure}
    
    \caption{Examples from \spotedit~\textit{standard} section.}
    \label{fig:more-standard_category}
\end{figure}

\begin{figure}[H]
    \centering
    \begin{subfigure}[b]{1.0\linewidth}
        \includegraphics[width=\linewidth]{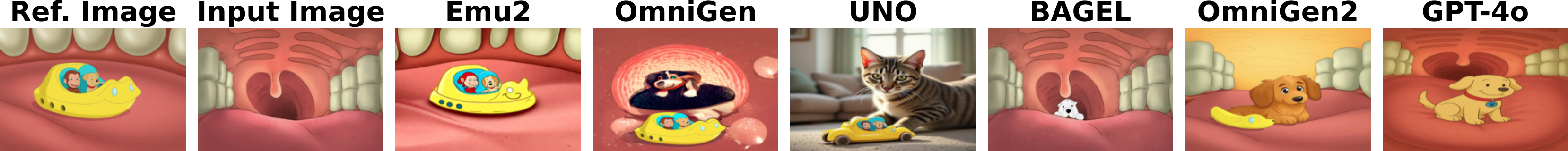}
        \caption{\textit{Instruction}: Replace the striped cat in the second image with the dog in the first image.}
        \label{fig:app_rob_syn_0}
    \end{subfigure}\\
    \begin{subfigure}[b]{1.0\linewidth}
        \includegraphics[width=\linewidth]{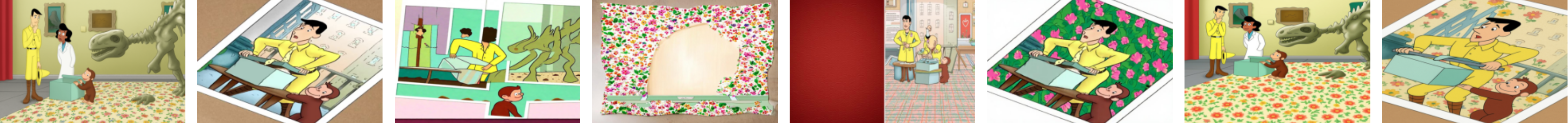}
        \caption{\textit{Instruction}: Replace the red surface in the second image with a flowery surface in the first image.}
        \label{fig:app_rob_syn_2}
    \end{subfigure}\\
    \begin{subfigure}[b]{1.0\linewidth}
        \includegraphics[width=\linewidth]{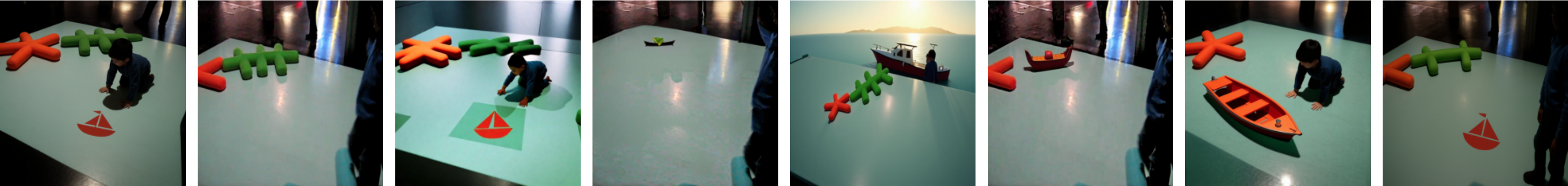}
        \caption{\textit{Instruction}: Replace the green fish in the second image with the boat in the first image.}
        \label{fig:app_rob_real_2}
    \end{subfigure}
    \caption{Examples from \spotedit~\hallucination~section.}
    \label{fig:more-rob_category}
\end{figure}

\paragraph{Additional qualitative examples.} We provide further qualitative examples from the \spotedit~benchmark. Fig.~\ref{fig:more-standard_category} presents additional cases from the standard section, while Fig.~\ref{fig:more-rob_category} showcases representative samples from the \hallucination~section of the benchmark.

\section{Evaluations Details}
\label{app:evaluation_details}
\subsection{Evaluation metrics}
\begin{itemize}[itemsep=0mm, leftmargin=*]
    \item \textbf{\oscore}: As discussed in Section~\ref{sec:spotedit}, each benchmark sample comprises four components: a reference image (providing visual guidance), a source image (to be edited), a textual prompt, and a ground-truth edited frame as the target output. The target outputs are first generated using GPT-4o and then refined through human supervision to ensure accuracy. To compute the \oscore, we measure the similarity between the edited image produced by baseline models and the ground-truth target output. It is important to note that \oscore~is the only metric that relies on the ground-truth annotations; all subsequent metrics depend solely on the reference and input images for evaluation.
    \item \textbf{\backscore}: As a more fine-grained evaluation, we assess the model’s ability to preserve the background of the input image while performing the required edit and generating the edited (or output) image. To conduct this evaluation, we first employ GroundingDINO~\citep{liu2023grounding} to generate bounding boxes around the source object in the input image and the target object in the output image. We then mask out these bounding boxes and compute the similarity score exclusively on the remaining background regions.
    \item \textbf{\objscore}: As a complementary evaluation to \backscore, we assess how well the model preserves the identity and appearance of the target object while following the visual guidance from the reference image and applying the edit to the input image. To this end, we employ GroundingDINO~\citep{liu2023grounding} to extract the target object from both the reference image and the output image. Once isolated, we apply a similarity metric between the two cut-out objects to evaluate how closely they align.
    \item \textbf{\fscore}: This metric specifically evaluates hallucination-induced failures. As discussed in Section~\ref{sec:hall-eval}, for both \irobscore~and \rrobscore~samples, the expected behavior is that the output image should remain identical to the input image, meaning that the target object must not be added during editing. To assess whether models avoid such unintended modifications, we employ the multimodal LLM InternVL3-8B as a binary classifier to determine whether the target object appears in the edited image. Since this is a binary classification task, the \fscore~is reported as accuracy.
\end{itemize}

\subsection{Comparison to DreamEdit} 
As discussed in Section~\ref{sec:related-work} and shown in Fig.\ref{fig:early-benchmarks}, prior benchmarks typically involve simple scenes, few distractors, and nearly identical object poses across the reference and edited images. To quantitatively assess the increased difficulty of our benchmark compared to the previously proposed DreamEdit benchmark\citep{li2023dreamedit}, we evaluate models on DreamEdit’s object replacement task, which consists of 198 samples. We use these samples to generate edited images with baseline models and then evaluate their performance using the metrics \oscore, \backscore, and \objscore.

For image similarity computation, we employ the \textsc{CLIP-ViT-H-14} model~\citep{Radford2021LearningTV, cherti2023reproducible} to extract image representations, given its strong performance across a wide range of semantic similarity tasks~\citep{ghazanfari2024towards}. Cosine similarity is then applied to the representations, yielding scores in the range $0 \leq \text{score} \leq 1$.

The results are presented in Tab.~\ref{tab:comparison_scores}. As mentioned earlier, \spotedit~provides ground-truth outputs, enabling us to compute \oscore. In contrast, the DreamEdit benchmark lacks this property, and thus \oscore~cannot be measured. However, since \backscore~and \objscore~rely only on the reference and input images, they remain computable for DreamEdit. Across all models and metrics, except for a single case, scores on \spotedit~are consistently lower, indicating that it presents a more challenging and complex task than DreamEdit.

\begin{table}[H]
    \centering
    \tabcolsep=2pt
    \caption{DreamEdit benchmark complexity compared to our \spotedit~(our) benchmark.}
    \label{tab:comparison_scores}
    \vspace{2mm}
    \resizebox{\textwidth}{!}{%
    \begin{tabular}{L{20mm}|
    C{20mm}@{\hspace{5pt}}C{15mm}|
    C{20mm}@{\hspace{5pt}}C{20mm}|
    C{20mm}@{\hspace{5pt}}C{20mm}}
        \toprule
        \multirow{2}{*}{Model} & 
        \multicolumn{2}{c}{\textbf{\oscore}} & 
        \multicolumn{2}{c}{\textbf{\backscore}} & 
        \multicolumn{2}{c}{\textbf{\objscore}} \\
        \cmidrule{2-3} \cmidrule{4-5} \cmidrule{6-7}
        & DreamEdit & \spotedit~& DreamEdit & \spotedit~& DreamEdit & \spotedit~\\
        \midrule
        \textbf{Emu2}     & - & 0.639 & 0.679 & 0.587\textcolor{Maroon}{\,$\downarrow$0.09} & 0.643 & 0.575\textcolor{Maroon}{\,$\downarrow$0.07} \\
        \textbf{OmniGen}  & - & 0.429 & 0.730 & 0.497\textcolor{Maroon}{\,$\downarrow$0.23} & 0.592 & 0.431\textcolor{Maroon}{\,$\downarrow$0.16} \\
        \textbf{UNO}      & - & 0.575 & 0.599 & 0.503\textcolor{Maroon}{\,$\downarrow$0.10} & 0.574 & 0.528\textcolor{Maroon}{\,$\downarrow$0.05} \\
        \textbf{BAGEL}    & - & 0.672 & 0.909 & 0.828\textcolor{Maroon}{\,$\downarrow$0.08} & 0.559 & 0.496\textcolor{Maroon}{\,$\downarrow$0.06} \\
        \textbf{OmniGen2} & - & 0.719 & 0.735 & 0.562\textcolor{Maroon}{\,$\downarrow$0.17} & 0.679 & 0.697\textcolor{OliveGreen}{\,$\uparrow$0.02} \\
        \bottomrule
    \end{tabular}}
\end{table}

\clearpage
\newpage

\end{document}

%% file: math_commands.tex
\usepackage{amsmath,amsfonts,bm}

\def\eqref#1{equation~\ref{#1}}

\def\1{\bm{1}}

\DeclareMathAlphabet{\mathsfit}{\encodingdefault}{\sfdefault}{m}{sl}
\SetMathAlphabet{\mathsfit}{bold}{\encodingdefault}{\sfdefault}{bx}{n}

